\documentclass[conference]{IEEEtran}
\IEEEoverridecommandlockouts
\usepackage{cite}
\usepackage{amsmath,amssymb,amsfonts}
\usepackage{algorithmic}
\usepackage{graphicx}
\usepackage{textcomp}
\usepackage{xcolor}
\def\BibTeX{{\rm B\kern-.05em{\sc i\kern-.025em b}\kern-.08em
    T\kern-.1667em\lower.7ex\hbox{E}\kern-.125emX}}
\usepackage{float}
\usepackage{url}

\usepackage{multirow}

\begin{document}

\title{CASC: Causal Adversarial Subspace Clustering for Multivariate Spatiotemporal Data\\}

\author{\IEEEauthorblockN{1\textsuperscript{st} Francis Ndikum Nji}
\IEEEauthorblockA{\textit{Dept. of Information Systems} \\
\textit{University of Maryland, Baltimore County}\\
Baltimore, MD, U.S.A \\
fnji1@umbc.edu}
\and
\IEEEauthorblockN{2\textsuperscript{nd} Vandana Janeja}
\IEEEauthorblockA{\textit{Dept. of Information Systems} \\
\textit{University of Maryland, Baltimore County}\\
Baltimore, MD, U.S.A \\
vjaneja@umbc.edu}
\and
\IEEEauthorblockN{3\textsuperscript{rd} Jianwu Wang}
\IEEEauthorblockA{\textit{Dept. of Information Systems} \\
\textit{University of Maryland, Baltimore County}\\
Baltimore, MD, U.S.A \\
jianwu@umbc.edu}
}

\maketitle
\begin{abstract} Deep subspace clustering plays a critical role in applications involving multivariate spatiotemporal data, such as sea ice monitoring, disease spread analysis, and tracking neuro-degeneration over time. Despite recent advances, existing methods primarily rely on geometric self-expressiveness, assume static subspace structures, and often fail to capture causal dependencies, local spatial interactions, and long-range temporal dynamics inherent in complex spatiotemporal systems. To address these limitations, we propose a novel \textbf{Causal Adversarial Subspace Clustering (CASC)} framework for discovering evolving latent regimes in high-dimensional spatiotemporal data. CASC integrates a U-Net-inspired deep adversarial clustering architecture with stacked FAConvLSTM layers to preserve spatial and temporal structure while learning robust latent representations. A graph attention transformer-based self-expressive network is introduced to jointly model local spatial relationships, global dependencies, and long-range temporal interactions. Furthermore, we propose two new learning objectives: (1) a \textit{Causal Subspace Preservation Loss} that aligns self-expression coefficients with latent causal relationships, encouraging clusters to reflect underlying causal processes rather than simple feature similarity, and (2) a \textit{Dynamic Temporal Subspace Evolution Loss} that captures evolving subspace structures and temporal regime transitions in nonstationary environments. Together, these components transform deep subspace clustering from a correlation-driven paradigm into a causal-temporal regime discovery framework. Experiments on three real-world multivariate spatiotemporal datasets demonstrate that CASC consistently produces more compact, better-separated, and temporally coherent clusters than state-of-the-art deep clustering methods, yielding substantial improvements in clustering accuracy, interpretability, and robustness.

\end{abstract}

\begin{IEEEkeywords}
deep subspace clustering, generative adversarial networks, graph attention transformer, self expressive network, spatiotemporal data, causal discovery.
\end{IEEEkeywords}
\vspace{-0.5em}
\section{Introduction}\label{sec:intro}
Recent years have seen increased availability in spatiotemporal data from common sources such as government surveys, mobile and wearable devices, launched satellite and weather sensors. These  data sources acquire, compress, store, transmit, and process massive amounts of complex high-dimensional multivariate spatiotemporal data. Although this data is high dimensional~\cite{yang2021investigating}, their intrinsic dimension (i.e number of variables needed to describe a data distribution) is often much smaller than the dimension of the ambient space~\cite{pope2021intrinsic}. For example; in image data processing, the number of pixels in an image can be rather large, yet most image processing models use only a few parameters to describe, for instance the appearance, geometry, and dynamics of a scene. This notion has motivated the development of techniques like autoencoders and regularization methods~\cite{gonzalez2018deep, zhu2018ldmnet} for representing high-dimensional data in a lower dimension. Another technique for representing high-dimensional dataset in a lower dimension is the Principal Component Analysis (PCA) \cite{kurita2019principal}. It assumes that the data is drawn from a single low-dimensional subspace within a high-dimensional space. However, in practice, data points may come from multiple subspaces, and the membership of these points to their respective subspaces is often unknown. This creates a complex sample distribution problem, particularly in multidimensional spatiotemporal data. Therefore, it is necessary to group data points into clusters, where each cluster contains points from the same subspace. This grouping assumes that data lies in different subspaces~\cite{chen2020stochastic}. A category of classical subspace clustering methods have been proposed~\cite{chen2020stochastic, liu2012robust,xu2021selective, ding2024survey}. Recent advances in deep subspace clustering~\cite{yang2019deep,dang2020multi,li2021lrsc,ji2017deep} have shown that combining representation learning with subspace discovery significantly improves clustering performance. Despite their success~\cite{zhang2018scalable,fan2021large,zhang2021learning}, existing deep subspace clustering methods suffer from several important limitations. First, they predominantly rely on geometric similarity and self-expression while overlooking causal relationships that govern interactions within complex spatiotemporal systems. Second, they generally assume static subspace structures, despite the fact that many real-world phenomena evolve continuously over time. Third, they often fail to jointly capture local spatial dependencies, long-range temporal interactions, and positional information necessary for understanding complex multiscale processes. Finally, the application of deep subspace clustering to multivariate spatiotemporal data remains relatively underexplored.

To address these challenges, we propose a novel Causal Adversarial Subspace Clustering (CASC) framework. The proposed architecture consists of a deep subspace clustering generator and a quality-verifying adversarial discriminator that are trained jointly in an end-to-end manner. Inspired by the success of U-Net architectures in representation learning \cite{ronneberger2015u}, the generator employs stacked TimeDistributed FAConvLSTM \cite{nji2026faconvlstm} layers to preserve spatial and temporal structure while learning compact latent representations. A graph attention transformer-based self-expressive network is introduced at the bottleneck to simultaneously capture local spatial interactions, global dependencies, positional awareness, and long-range temporal correlations. More importantly, we extend conventional deep subspace clustering by introducing two novel learning mechanisms. First, we propose a \textit{Causal Subspace Preservation Loss}, which explicitly aligns the learned self-expression structure with latent causal relationships. This encourages the discovered clusters to reflect underlying causal processes rather than simple feature similarity. Second, we introduce a \textit{Dynamic Temporal Subspace Evolution Loss} that enables the model to learn evolving subspace representations and capture temporal regime transitions in nonstationary environments. Together, these components transform deep subspace clustering from a correlation-driven paradigm into a causal-temporal regime discovery framework. The proposed framework further incorporates a clustering layer based on Student's t-distribution to iteratively refine cluster assignments and improve latent space organization. Simultaneously, the decoder reconstructs the original multidimensional spatiotemporal observations, ensuring that latent representations preserve both semantic and structural information. A subspace-aware energy-based temporal discriminator evaluates the quality of learned representations and guides the generator toward producing more discriminative and cluster-friendly embeddings. To sum
up, this paper makes the following contributions: 1) We propose a novel Causal Adversarial Subspace Clustering (CASC) framework for high-dimensional multivariate spatiotemporal data that jointly performs representation learning, adversarial learning, and subspace clustering in an end-to-end manner. 2) We introduce a Causal Subspace Preservation Loss that incorporates latent causal relationships into the self-expression learning process, enabling the discovery of causally coherent subspaces and improving clustering interpretability. 3) We propose a Dynamic Temporal Subspace Evolution Loss that models evolving subspace structures over time and captures temporal regime transitions in nonstationary spatiotemporal systems. 4) We propose a novel Subspace-Aware Energy-based Temporal Discriminator that directly measures how well latent representations conform to cluster-specific subspaces.

The remainder of the paper is structured as follows. Section \ref{bknrw} summarizes the background while  \ref{sec:related} discusses related works. Section \ref{sec:prob_st} describes the problem in detail while Section \ref{sec:method} presents our proposed solution. Section \ref{sec:exp} details the experiment and presents our results while Section \ref{concl} concludes our research.
\vspace{-0.6em}
\section{Background and Motivation} \label{bknrw}
\vspace{-0.6em}
The exponential growth of multivariate spatiotemporal data across disciplines has created both unprecedented opportunities and formidable analytical challenges. These data are high-dimensional, noisy, heterogeneous, and often exhibit strong nonlinear dependencies across space, time, and variables. Conventional clustering methods, which treat samples as independent and identically distributed (iid), fail to capture these intricate dependencies and often miss the latent low-dimensional subspace structure that governs real-world dynamics. This motivates the pursuit of deep subspace clustering (DSC) methods that uncover meaningful representations from complex spatiotemporal systems, disentangle overlapping patterns, and group data into coherent clusters that are physically interpretable and temporally consistent.
For multivariate spatiotemporal data, these subspaces may represent distinct climate regimes, transportation flow patterns, disease outbreak waves, or other structured phenomena. The development of deep neural architectures particularly those leveraging convolutional, recurrent, and attention-based modules enable learning hierarchical feature representations that preserve spatial locality, model temporal continuity, and capture complex cross-variable correlations. Integrating subspace clustering with representation learning is therefore a powerful paradigm: it simultaneously discovers latent feature space and can segment data into meaningful subspaces, improving robustness to noise and scalability to large datasets.

The motivation for this research is also deeply societal. For instance, in climate science, accurately clustering snowmelt regions, sea-ice zones, or drought-affected areas can improve predictions of sea-level rise, inform resource allocation for adaptation, and guide early warning systems for vulnerable communities. In epidemiology, spatiotemporal clustering can reveal emerging hotspots of disease transmission and support timely interventions. Developing robust, interpretable, and generalizable deep subspace clustering models thus contribute not only to advancing deep learning theories but also to support decisions in high-stakes domains where timely insights can save lives, protect infrastructure, and shape policy.

\section{Related Work}\label{sec:related}

\textbf{Self-expressive learning for deep subspace clustering}.
These methods learn self-expression coefficient matrices that capture the relationships between data points. Given a data matrix \( X \in \mathbb{R}^{d \times n} \), we express each data point as a linear combination of other data points as: \(X = X M\), where \( M \in \mathbb{R}^{n \times n} \) is the self-expression coefficient matrix. The optimization problem is: \(\min_{M} \| X - X M \|_F^2 + \lambda \|M\|_1\). Inspired by recent advances in deep learning, Zhang et al., \cite{zhang2021learning} proposed a novel framework for subspace clustering Self-Expressive Network (SENet), which employs two multilayer preceptrons (MLPs) referred to as Query-Net and Key-net to learn a self-expressive representation of the data. While SENet may work well on out of sample data, it struggles to capture long-range dependencies and positional awareness, a vital component for subspace clustering of multivariate spatiotemporal data. Recently, Baek et al., \cite{baek2021deep} proposed Deep self-representative subspace clustering network for unsupervised subspace clustering to improve representativeness and clustering ability. 
Although they attempt to improve clustering ability, they completely rely on self expression as supervision and do not preserve local features or geometric relationships between data point. Recently Zhao et al., \cite{zhao2023deep} proposed a double self-expressive subspace clustering algorithm which improves performance by preserving the structural information in the self-expressive coefficient matrix.

\textbf{Adversarial Networks for subspace clustering.}
Recently, there is growing interests in combining the strengths of GANs with subspace clustering methods to enhance clustering performance in complex high-dimensional datasets. Zhou et al. proposed a Deep Adversarial Subspace Clustering (DASC) \cite{zhou2018deep} which introduces adversarial learning and supervises the generator’s learning to produce more favorable representations for better subspace clustering. While they addressed the clustering error with little reliance on self-expression for supervision, they overlooked local features, useful long-range dependencies and positional information in feature representation. Mukherjee et al., \cite{mukherjee2019clustergan} proposed clusterGAN and demonstrated that while one can potentially exploit the latent-space back-projection in GANs to cluster, the cluster structure is not retained in the GAN latent space. Recently, Yu et al., \cite{yu2020gan} proposed two GAN-based enhanced deep subspace clustering approaches: deep subspace clustering via dual adversarial generative networks (DSC-DAG) and self-supervised deep subspace clustering with adversarial generative networks (\(S^2\) DSC-AG) and use adversarial training to simultaneously learn the distributions of both the inputs and latent representations.

\textbf{Deep Learning based Clustering.} The limitations of traditional clustering methods have motivated the development of deep learning-based approaches, which are better equipped to model nonlinear, high-dimensional data. Deep Embedded Clustering (DEC) \cite{xie2016unsupervised} introduced the paradigm of jointly learning representations and cluster assignments by minimizing a Kullback–Leibler (KL) divergence loss between predicted and target distributions. Extensions of DEC and related autoencoder-based methods have been applied to time series data, though many approaches either focus solely on temporal patterns or image-level spatial similarities, neglecting the joint spatiotemporal structure. To address this gap, spatiotemporal autoencoders \cite{faruque2023deep} have emerged, combining convolutional neural networks (CNNs) with recurrent architectures such as Long Short-Term Memory (LSTM) networks, proving highly effective for spatiotemporal data, as it integrates convolutional operations into recurrent units, allowing the model to simultaneously capture localized spatial patterns and their evolution over time. This approach has been successfully applied to applicatins such as precipitation nowcasting \cite{shi2017deep}, sea ice prediction in the Arctic \cite{wang2019deep}, and regional climate variability detection \cite{liu2020spatiotemporal}, demonstrating its capability to extract meaningful representations from complex spatiotemporal climate datasets. These applications highlight the model’s ability to capture both fine-scale spatial temporal dependencies. 

\section{Problem Definition}\label{sec:prob_st}

Let \( U = \bigcup_{i=1}^{M} S_i \) be the nonlinear set consisting of a union of \( M \) subspaces \( \{S_i \subset H\}_{i=1}^{M} \), where \( S_i \) are subspaces of a Hilbert or a Banach space \( H \). Let \( W = \{w_j \in H\}_{j=1}^{N} \) be a set of multivariate spatiotemporal data points drawn from \( U \). Each data point \( w_j \) is represented as a high-dimensional tensor \( w_j \in \mathbb{R}^{T \times \text{lon} \times \text{lat} \times \text{var}} \), where \( T \) denotes the temporal dimension, and \( \textit{lon}, \textit{lat} \) denote the spatial dimensions across multiple variables \( \textit{var} \). \\
\textbf{Goal:} Our objective is to segment this temporal sequence into $K$ coherent clusters 
$\{\mathcal{C}_1, \mathcal{C}_2, \ldots, \mathcal{C}_K\}$ along the temporal dimension such that time steps within the same cluster share similar 
\emph{latent subspace representations} that capture  both their temporal structure and multivariate interactions.
For a predefined number of subspaces $M$ with intrinsic dimensions $\{d_i\}_{i=1}^{M}$, the problem can be formalized as the following minimization: \(e(W, S) := \sum_{f \in W} \min_{1 \leq j \leq M} d_{\mathcal{H}}(f, S_j),\) where $S = \{S_1, \dots, S_M\}$ is a candidate set of subspaces, $d_{\mathcal{H}}(\cdot, \cdot)$ is the distance induced by the norm on $\mathcal{H}$, and $e(W,S)$ measures the total reconstruction error.  
The task is to find \(S^* = \{ S_1^*, \dots, S_M^* \} = \arg \min_{S \in \mathcal{S}} e(W, S).\)\\
\textit{Learning Orthonormal Bases.} For each subspace $S_i$, we seek an orthonormal basis \(\{u_{i1}, \dots, u_{i d_i}\} \subset S_i\), such that \(S_i = \mathrm{span}\{u_{i1}, \dots, u_{i d_i}\}, \quad \langle u_{ij}, u_{ik} \rangle_{\mathcal{H}} = \delta_{jk},\)
where $d_i = \dim(S_i) \ll \dim(\mathcal{H})$ is the intrinsic dimension, $\langle \cdot, \cdot \rangle_{\mathcal{H}}$ is the inner product in $\mathcal{H}$, and $\delta_{jk}$ is the Kronecker delta. In deep subspace clustering, these basis vectors are learned implicitly via a deep encoder $f_{\theta}$ producing latent representations: \(z_j = f_{\theta}(w_j) \in \mathbb{R}^{d}.\) Points from the same subspace are expected to lie near a linear subspace in $\mathbb{R}^{d}$, allowing PCA or SVD to recover an orthonormal basis for each subspace.
\textit{Clustering Assignment: }Define a clustering function \(\varphi: W \to \{1, \dots, M\},\)
such that each $w_j \in W$ is assigned to a unique subspace $S_{\varphi(w_j)}$.  
The resulting clusters are \(C_i = \{ w_j \in W \mid \varphi(w_j) = i \}, \quad i = 1, \dots, M.\)
\textit{Clustering Constraints.} The clusters must satisfy: 1) \textit{Partition:} $\bigcup_{i=1}^{M} C_i = W$, 2). \textit{Disjointness:} $C_i \cap C_j = \emptyset, \quad \forall i \neq j$, and 3) \textit{Subspace Membership:} $C_i \subset S_i \subset \mathcal{H}$ for each $i$.

\section{Methodology}\label{sec:method}
In this section, we present the architecture and training strategy of the proposed Causal Adversarial Subspace Clustering (CASC) framework. As shown in Figure \ref{fig:j}, CASC couples a spatiotemporal encoder-decoder with a bidirectional temporal graph attention bottleneck, a DEC-style per-timestep clustering head with temperature and balancing regularizers, a self-expressive temporal layer in the generator \(G\), and a Subspace-Aware Energy-based Temporal Discriminator  \(D\).
\begin{figure*}[ht!]
    \centering
    \includegraphics[width=0.9\textwidth]{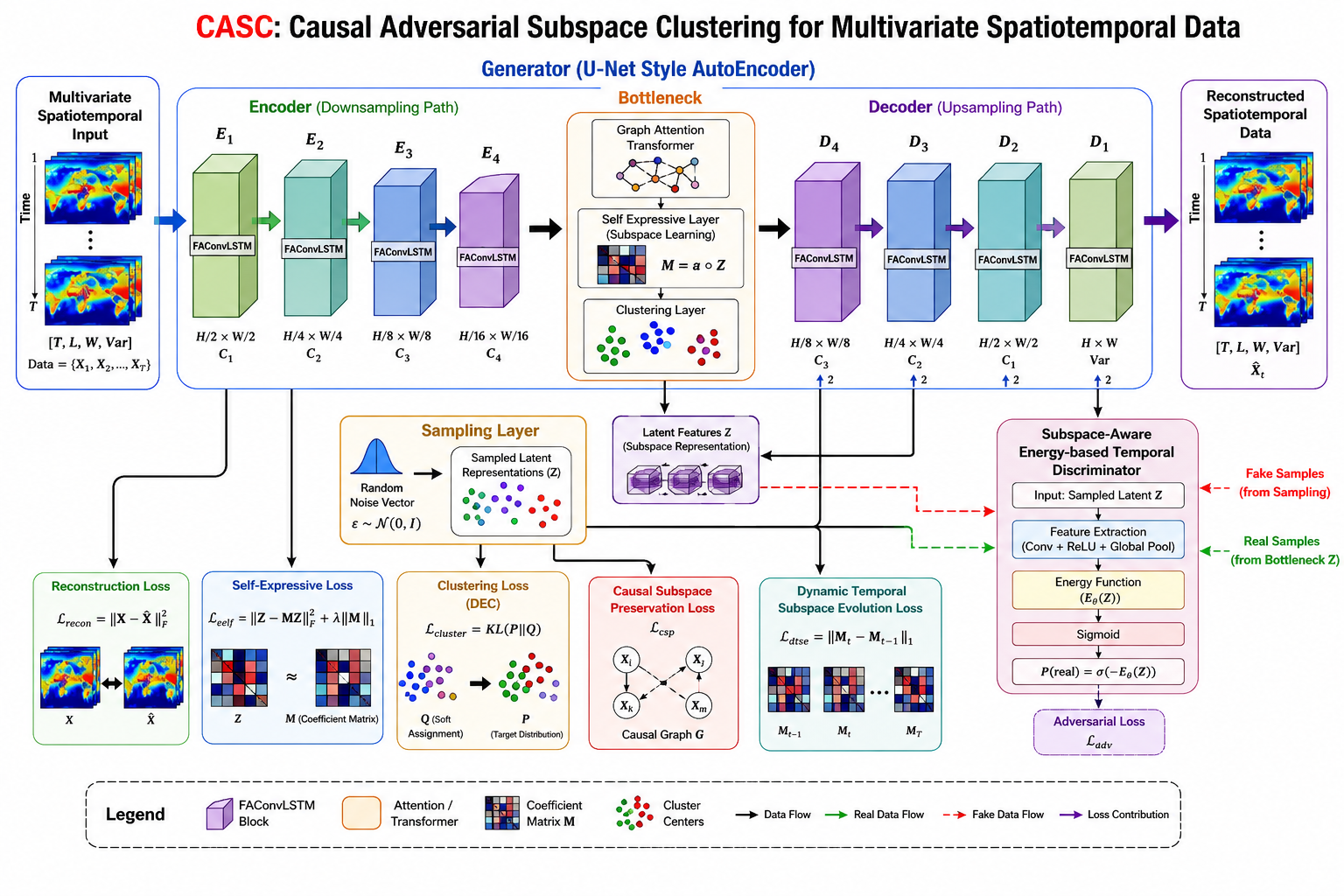}
    \caption{{Architecture of Causal Adversarial Subspace Clustering for Multivariate Spatiotemporal Data (CASC) framework.} CASC is composed of a deep subspace and clustering generator and a quality-verifying discriminator. The generator is composed of an autoencoder and a sampling layer. The autoencoder receives 4D multivariate spatiotemporal data as input and outputs cluster labels, self-expressive coefficient matrix and reconstructed data. The sampling layer receives as input the cluster labels, coefficient matrix and random noise vectors and outputs real clustered data features and fake data features. Both feature vectors are sent to the discriminator to determine real or fake data features while generating new data.}
    \label{fig:j}
\end{figure*}

\subsection{Generator (\(G\))}
\(G\) is designed to jointly learn hierarchical spatiotemporal representations, a causally informed temporal affinity structure, and a clustering assignment that is robust, balanced, and interpretable. \(G\) is composed of four key components: (i) a FAConvLSTM-based spatiotemporal encoder, (ii) a bidirectional temporal graph attention transformer (Bi-TGAT) layer, (iii) a self-expressive temporal subspace layer, and (iv) a decoder with skip connections to ensure faithful reconstruction. Figure~\ref{fig:j} provides a schematic of the end-to-end pipeline.

\subsubsection{Notation and Input Representation}Let the input be a multivariate spatiotemporal tensor
\(\mathbf{X} \in \mathbb{R}^{B \times T \times H \times W \times C},\) with batch size $B$, time steps $T$, spatial grid $H\times W$, and $C$ variables. We write $\mathbf{X}_b$ for the $b$-th sample, and $\mathbf{X}_{t,:,:,:} \in \mathbb{R}^{H \times W \times C}$ for the $t$-th frame of a sequence. Our goal is to produce per-timestep soft cluster assignments $\mathbf{q}_{t} \in \Delta^{K-1}$ over $K$ clusters and to learn subspace-aware embeddings $\mathbf{z}_{t}\in\mathbb{R}^{D}$ that are discriminative, temporally coherent, and subspace-preserving.
\subsubsection{Spatiotemporal Encoder (FAConvLSTM with Residual Temporal Blocks)}The encoding phase follows a U-Net downsampling structure composed of five FAConvLSTM layers. This encoder is responsible for learning low-dimensional latent representations that capture both spatial and temporal correlations. By applying convolutions across both space and time, the encoder compresses the input data into a bottleneck representation 
\( Z \in \mathbb{R}^{T \times d} \), where \( T \) denotes the temporal dimension and \( d \) the latent feature dimensionality. For brevity, let \( \{X_1 \dots , X_n\}\) denote the input samples and let \( \{z_1 \dots , z_n\}\) denote their corresponding latent representations learned by the encoder in \(G\). Namely, \(z _i \in \mathbb{R}^d\) is the d-dimensional representation of the \(i-\)th 4D sample \(X_i \in \mathbb{R}^{T \times H \times W \times C}\) and \(k\) denotes the number of clusters or subspaces. The mapping from an input data space to the latent feature space is a nonlinear function \(f_{\text{enc}}:= X \to Z\), were \(Z \in \mathbb{R}^m\) is an \(m\)-dimensional high-level representation of all the variables at each timestep. \(Z_t = \) \(f_{\text{enc}} (X_t)\), \(t = \{1, \dots , T\}\). From Figure \ref{fig:j}, to extract spatial, temporal and salient features at different scales and reduced dimensionality, the encoder applies a FAConvLSTM stem followed by residual temporal blocks and temporal-preserving spatial pooling: 
\begin{align}
\mathbf{H}^{(1)}&=\mathrm{FAConvLSTM}_{64}(\mathbf{X}),\\
\mathbf{H}^{(l)}&=\mathrm{ResTempBlock}_{F_l}(\mathrm{MaxPool3D}_{(1,2,2)}(\mathbf{H}^{(l-1)}))
\end{align}
for $l=2,3,4$ with filters $F_l\in\{64,128, 256, 512, 1024\}$. At encoder level $l$, the input sequence is processed by an FAConvLSTM cell that captures both temporal dependencies and spatial correlations. For each time step $t$, the hidden state $\mathbf{h}_t^{(l)}$ and memory state $\mathbf{c}_t^{(l)}$ are updated according to \((\mathbf{h}_t^{(l)},\mathbf{c}_t^{(l)})
=
\mathrm{FAConvLSTM}
\left(
\mathbf{x}_t^{(l)},
\mathbf{h}_{t-1}^{(l)},
\mathbf{c}_{t-1}^{(l)}
\right),\) where $\mathbf{x}_t^{(l)}$ denotes the spatial feature map at time step $t$.

Unlike conventional CNN encoders that process frames independently, the FAConvLSTM layer propagates hidden states through time, enabling the network to learn long-range temporal interactions while preserving local spatial structures. The outputs from all time steps are stacked to form \(\mathbf{H}^{(l)}
=
[\mathbf{h}_1^{(l)},\mathbf{h}_2^{(l)},\ldots,\mathbf{h}_T^{(l)}].\) The final hidden representation from each encoder stage; \(\mathbf{S}^{(l)}=\mathbf{h}_T^{(l)},\) is preserved as a skip connection and forwarded to the decoder.

To construct a hierarchical feature pyramid, a time-distributed convolutional downsampling operation is applied after each FAConvLSTM block. Specifically, \(\mathbf{H}_{down}^{(l)}
=
\mathrm{Conv}_{3\times3}^{s=2}
\left(
\mathbf{H}^{(l)}
\right),\) where the stride of two reduces the spatial resolution while increasing the representational capacity. Repeating this procedure across five encoder stages progressively transforms the original sequence into a compact latent spatiotemporal representation.

\subsubsection{Latent Bottleneck Representation}

After the final encoder stage, the compressed latent sequence is represented as \(\mathbf{Z}
\in
\mathbb{R}^{B \times T \times H_b \times W_b \times 1024},\) where $H_b$ and $W_b$ denote the bottleneck spatial dimensions. The bottleneck aggregates high-level temporal dynamics and multi-scale spatial information extracted throughout the encoder hierarchy.

\subsubsection{Patchification for Graph Efficiency. }To form a tractable spatiotemporal node set for attention, we tile the top-level tensor into non-overlapping patches of size $(h_p, w_p)$, yielding a reduced grid $(H',W')$ with $N{=}H'W'$ nodes per frame. We flatten $(H',W')$ to a node axis so that the sequence becomes 
\(
\mathbf{X}^{(4)} \rightarrow \tilde{\mathbf{X}} \in \mathbb{R}^{B\times T \times N \times F},
\)
with $F{=}1024$. Patchification reduces graph size and stabilizes attention training while preserving local spatial structure.

\subsubsection{Bidirectional Temporal Graph Attention (Bi-TGAT) Bottleneck}We process the node sequences with a bidirectional temporal GAT layer that aggregates information both forward and backward in time. Let $\mathbf{H}_t\in\mathbb{R}^{N\times F}$ be node features at time $t$. A temporal adjacency (implicit, learned) is built by attention over $\mathbf{H}_t$ and $\mathbf{H}_{t\pm1}$; for each direction $d\in\{\rightarrow,\leftarrow\}$ we compute:
\vspace{-0.5em}
\begin{equation}
     \alpha^{(d)}_{t,i\to j} \&= \text{softmax}_j\left(\phi\big(\mathbf{W}^{(d)}_q \mathbf{h}_{t,i},\, \mathbf{W}^{(d)}_k \mathbf{h}_{t+\delta_d,j}\big)\right)
\end{equation}
where \(\delta_{\rightarrow}{=}+1,\ \delta_{\leftarrow}{=}-1,\)
\vspace{-0.9em}
\begin{equation}
    \mathbf{m}^{(d)}_{t,i} \&= \sum_{j} \alpha^{(d)}_{t,i\to j}\, \mathbf{W}^{(d)}_v \mathbf{h}_{t+\delta_d,j},
\end{equation} where \(\mathbf{h}'_{t,i} = \text{LN}\!\left(\mathbf{W}_o [\mathbf{m}^{(\rightarrow)}_{t,i} \Vert \mathbf{m}^{(\leftarrow)}_{t,i}] \right).\)

Multi-head attention stabilizes learning; the outputs are pooled across $N$ nodes to obtain a per-timestep embedding $\mathbf{z}_t\in\mathbb{R}^{D}$ (via head concat + projection), and globally pooled across $t$ to get a sequence summary $\bar{\mathbf{z}} \in \mathbb{R}^{D}$. A linear projection and LayerNorm yield the final temporal sequence embeddings $\{\mathbf{z}_t\}_{t=1}^T$. By integrating Bi-TGAT, we fuse local spatial context with directed temporal cues ($t\pm1$), mitigating exposure bias and enhancing discriminability of transient regimes. Attention weights act as data-adaptive temporal edges, improving subspace separation by emphasizing causally or dynamically influential frames.

\subsubsection{Per-Timestep Clustering Head with Temperature and Balance}
We adopt a DEC-style Student-$t$ assignment per time step~\cite{xie2016unsupervised}:
\vspace{-0.9em}
\begin{align}
q_{t,k} \propto \left(1 + \frac{\|\mathbf{z}_t - \boldsymbol{\mu}_k\|_2^2}{\alpha}\right)^{-\frac{\alpha+1}{2}},
\quad
\sum_{k=1}^{K} q_{t,k}=1,
\end{align}
with trainable centers $\{\boldsymbol{\mu}_k\}_{k=1}^K$ and dof $\alpha$. We introduce a temperature $\tau$ to control sharpness:
\(
\tilde{q}_{t,k} \propto q_{t,k}^{1/\tau},
\)
annealing $\tau \!\downarrow$ to harden assignments over training. To avoid mode collapse, we combine two balancing terms: (i) a batch-wise KL divergence to a near-uniform marginal to enforce cluster utilization, and (ii) a mutual-information style redundancy penalty across clusters. The head optimizes the classic DEC target distribution $\mathbf{p}$ computed from $\mathbf{q}$ and adds $\mathrm{KL}(\mathbf{p}\|\mathbf{q})$ to the loss. DEC sharpening aligns the extracted features $\mathbf{z}_t$ with centers, while temperature scheduling prevents early overcommitment. Balanced assignments keep clusters populated and reduce mode collapse.

\subsubsection{Causal Subspace Preservation Loss}
\label{subsec:CSP}
Traditional deep subspace clustering relies on the self-expressiveness property, which assumes that each latent representation can be reconstructed as a linear combination of other latent representations belonging to the same subspace. 
Let \(\mathbf{Z}=[z_1,z_2,\ldots,z_N] \in \mathbb{R}^{d \times N}\) denote the latent representations learned by the encoder, where $N$ is the number of samples and $d$ is the latent dimensionality. The conventional self-expression objective is given by: \(L_{\text{self}}=
\left\|
\mathbf{Z}-\mathbf{Z}\mathbf{C}
\right\|_F^2
+
\lambda_1 \|\mathbf{C}\|_p,\) where $\mathbf{C}\in\mathbb{R}^{N\times N}$ denotes the self-expression coefficient matrix and $\|\cdot\|_p$ is either an $\ell_1$ or $\ell_2$ regularization term. Although effective, this formulation only preserves geometric similarity and ignores the underlying causal interactions governing complex spatiotemporal systems. In climate data, for example, atmospheric circulation, ocean-atmosphere coupling, and teleconnection patterns often drive observed variability. Consequently, samples belonging to the same causal process may not necessarily exhibit strong geometric similarity in the latent space.

To address this limitation, we introduce a \textit{Causal Subspace Preservation Loss (CSP)} that explicitly aligns the learned subspace structure with latent causal dependencies. Let \(\mathbf{A}_{\text{causal}}
\in
\mathbb{R}^{N\times N}\) denote a causal adjacency matrix estimated from latent features using a causal discovery mechanism such as Neural Granger Causality, PCMCI+, Transfer Entropy, or Convergent Cross Mapping. The affinity structure induced by the self-expression coefficients is represented by \(\mathbf{S}
=
\mathbf{C}\mathbf{C}^{T}.\) The proposed causal consistency objective is defined as:
\vspace{-0.5em}
\begin{equation}
L_{\text{causal}}
=
\left\|
\mathbf{A}_{\text{causal}}
-
\mathbf{C}\mathbf{C}^{T}
\right\|_F^2.
\end{equation} The complete CSP loss becomes
\begin{equation}
L_{\text{CSP}}
=
\left\|
\mathbf{Z}-\mathbf{Z}\mathbf{C}
\right\|_F^2
+
\lambda_c
\left\|
\mathbf{A}_{\text{causal}}
-
\mathbf{C}\mathbf{C}^{T}
\right\|_F^2
+
\lambda_r
\|\mathbf{C}\|_p.
\end{equation}

The first term preserves the classical self-expressiveness property, while the second term encourages the learned subspaces to respect causal relationships discovered from the data. As a result, samples belonging to the same underlying physical process are encouraged to reside within the same subspace. This produces clusters that are not only geometrically coherent but causally meaningful and scientifically interpretable.
From a climate science perspective, the CSP loss allows the model to identify latent climate regimes associated with common driving mechanisms rather than merely grouping observations based on similarity. Consequently, it improves cluster compactness, separation, robustness, and interpretability.
\subsubsection{Dynamic Temporal Subspace Evolution Loss}
\label{subsec:DTSE}

Most existing deep subspace clustering methods assume that the underlying subspace structure remains fixed throughout the observation period. This assumption is often unrealistic for multivariate spatiotemporal data because many real-world systems evolve continuously over time. Examples include seasonal transitions, snow accumulation and melting cycles, sea-ice formation and breakup, disease outbreaks, and traffic congestion patterns.
To model these evolving dynamics, we extend the conventional static self-expression matrix into a sequence of time-dependent self-expression matrices \(\mathcal{C}
=
\{
\mathbf{C}_1,
\mathbf{C}_2,
\ldots,
\mathbf{C}_T
\},\) where $\mathbf{C}_t$ represents the latent subspace structure at time step $t$ and $T$ denotes the total number of temporal observations. The self-expression reconstruction loss at time $t$ is given by \(L_{\text{self}}^{(t)}
=
\left\|
\mathbf{Z}_t
-
\mathbf{Z}_t \mathbf{C}_t
\right\|_F^2.\) To ensure temporal consistency while allowing subspaces to evolve naturally, we introduce a \textit{Dynamic Temporal Subspace Evolution (DTSE)} regularization term: \(L_{\text{DTSE}}
=
\sum_{t=2}^{T}
\left\|
\mathbf{C}_t
-
\mathbf{C}_{t-1}
\right\|_1.\) This formulation encourages neighboring temporal subspaces to remain similar while permitting gradual transitions over time. Alternatively, a probabilistic formulation can be employed using the Kullback-Leibler divergence: \(L_{\text{DTSE-KL}}
=
\sum_{t=2}^{T}
D_{\text{KL}}
(
\mathbf{C}_{t}
\|
\mathbf{C}_{t-1}
).\)
The complete dynamic subspace learning objective becomes:
\vspace{-0.9em}
\vspace{-0.5em}
\begin{equation}
L_{\text{dynamic}}
=
\sum_{t=1}^{T}
L_{\text{self}}^{(t)}
+
\lambda_d
L_{\text{DTSE}}.
\end{equation}

The DTSE loss provides several advantages. First, it captures evolving subspace structures and regime transitions that are common in nonstationary spatiotemporal systems. Second, it reduces abrupt cluster switching and improves temporal coherence. Third, it enables the model to identify persistent temporal regimes and transition boundaries. Finally, it improves the stability of cluster assignments across time while preserving the flexibility required to model complex dynamical processes.
By jointly optimizing the CSP and DTSE losses, the proposed framework discovers clusters that are simultaneously geometrically coherent, causally consistent, and temporally evolving. This transforms deep subspace clustering from a static correlation-based framework into a causal-temporal regime discovery framework suitable for complex multivariate spatiotemporal datasets.

\subsubsection{Decoder and Reconstruction Loss}
The decoder mirrors the encoder with ConvLSTM2D up-paths and skip connections from the encoder stages (temporal up-sampling aligns skip timings). The reconstruction loss
\(
\mathcal{L}_{\text{rec}} = \|\hat{\mathbf{X}} - \mathbf{X}\|_2^2
\)
anchors the representation to physically plausible spatiotemporal fields, improving stability of the latent space.


\subsubsection{The Sampling Layer} The generator additionally produces \emph{real} and \emph{fake} samples per cluster \(C_i\) 
using a reparameterization trick~\cite{kingma2013auto}:
\(
\bar{\mathbf{z}}_t = 
\sum_{j=1}^{m_i} \alpha_{tj} \mathbf{z}_{ij},
\quad t=1,\dots,m_i^{*},
\)
where \(\alpha_{tj}\sim \mathcal{U}(0,1]\) are fixed during training, 
allowing gradients to flow through \(\mathbf{z}_{ij}\). The generator is adversarially trained to \emph{reduce} fake residuals: \(\mathcal{L}_{\text{adv}} = \mathbb{E}_{\text{fake}}\,\mathcal{E}(\mathbf{z}_{\text{fake}};\mathbf{U}).\)
The discriminator shapes latent geometry to be union-of-subspaces: low residual within the correct cluster subspace and high residual otherwise. Orthogonality/separation regularizers improve cluster identifiability and reduce overlap between subspaces, increasing temporal cluster purity.

\vspace{-0.5em}
\subsection{Energy-Based Temporal Subspace Discriminator}
\label{subsec:discriminator} A key component of the proposed CASC framework is the \textit{Energy-Based Temporal Subspace Discriminator}, which serves as a quality verifier for the latent representations generated by the clustering network. Unlike conventional adversarial discriminators that learn a binary classification boundary between real and generated samples, the proposed discriminator evaluates how well a latent representation conforms to a set of learned subspace models. This design is particularly suitable for multivariate spatiotemporal clustering because the objective is not merely to distinguish real from fake samples, but rather to determine whether latent representations preserve the underlying subspace structure of the data.
Let \(\mathbf{z}_i \in \mathbb{R}^{d}\) denote a latent representation produced by the generator, where $d$ is the latent dimensionality. Instead of learning a deep neural classifier, the discriminator learns a collection of cluster-specific orthonormal subspace bases
\(\mathcal{U}
=
\{
\mathbf{U}_1,\mathbf{U}_2,\ldots,\mathbf{U}_K
\},\) where \(\mathbf{U}_k \in \mathbb{R}^{d \times r}\) represents the basis vectors of the $k$-th latent subspace, $K$ is the number of clusters, and $r$ is the subspace dimensionality.

For a given latent representation $\mathbf{z}$, the discriminator projects the sample onto each subspace: \(\hat{\mathbf{z}}_k
=
\mathbf{U}_k \mathbf{U}_k^{T}\mathbf{z},\) and computes the residual reconstruction error \(\mathbf{r}_k
=
\mathbf{z}
-
\hat{\mathbf{z}}_k.\) The corresponding subspace energy is defined as: \(E_k(\mathbf{z})
=
\|\mathbf{r}_k\|_2^2.\)

Intuitively, a sample that genuinely belongs to subspace $k$ should exhibit a small projection residual and therefore a low energy value. Conversely, samples that do not conform to the underlying subspace structure produce larger residuals and higher energies.

The discriminator assigns an overall energy score by selecting the minimum energy across all learned subspaces: \(E(\mathbf{z})
=
\min_{k}
E_k(\mathbf{z}).\) This formulation allows the discriminator to evaluate how well a latent representation aligns with the most compatible cluster-specific subspace.

\subsubsection{Adversarial Energy Objective}

Given real latent representations $\mathbf{z}^{real}$ and generated representations $\mathbf{z}^{fake}$, the discriminator is trained using a margin-based hinge loss: \(L_{hinge}
=
\frac{1}{N}
\sum_{i=1}^{N}
\max
\left(
0,
E_{real}^{(i)}
-
E_{fake}^{(i)}
+
m
\right),\) where $m$ denotes a predefined margin. This objective encourages real latent representations to exhibit lower energy than generated representations. G simultaneously minimizes \(L_G
=
\frac{1}{N}
\sum_{i=1}^{N}
E_{fake}^{(i)},\) thereby forcing generated latent representations to lie closer to the learned subspaces.

\subsubsection{Orthogonality Regularization}

To ensure that each subspace basis spans a valid low-dimensional manifold, orthogonality constraints are imposed: \(L_{orth}
=
\sum_{k=1}^{K}
\|
\mathbf{U}_k^{T}\mathbf{U}_k
-
\mathbf{I}
\|_F^2.\)

This regularization prevents basis collapse and promotes stable subspace learning.

\subsubsection{Subspace Separation Regularization}

To encourage cluster distinctiveness, the discriminator further minimizes overlap between different subspaces: \(L_{sep}
=
\sum_{i<j}
\|
\mathbf{U}_i^{T}\mathbf{U}_j
\|_F^2.\)
Minimizing this term promotes orthogonality between subspaces and increases cluster separability. The complete discriminator objective becomes \(L_D
=
L_{hinge}
+
\beta_{orth}L_{orth}
+
\beta_{sep}L_{sep}.\)
\subsubsection{Comparison with the DASC Discriminator}

The proposed discriminator differs fundamentally from the discriminator employed in DASC. The original DASC discriminator utilizes a deep neural architecture composed of multiple trainable layers that attempt to distinguish real and generated latent representations through classification. While effective, such discriminators introduce a large number of trainable parameters, increase memory consumption, and often suffer from adversarial instability. In contrast, the proposed Energy-Based Temporal Subspace Discriminator directly models the geometric structure of latent subspaces. Rather than learning complex nonlinear decision boundaries, it learns only a small set of low-dimensional subspace bases. Consequently, the number of trainable parameters scales as \(O(Kdr),\) whereas the complexity of a deep discriminator typically scales with multiple dense layers and nonlinear transformations.









Overall, the proposed Subspace-aware Energy-Based Temporal Discriminator provides a more efficient, stable alternative to the original DASC discriminator. By explicitly modeling cluster-specific subspaces and evaluating projection energies, it offers stronger alignment with the clustering objective while significantly reducing computational complexity.
\subsubsection{Overall Optimization Objective}

The complete CASC objective function is defined as:
\vspace{-0.5em}
\begin{equation}
\begin{aligned}
L_{\text{total}}
=
&
\lambda_1 L_{\text{rec}}
+
\lambda_2 L_{\text{adv}}
+
\lambda_3 L_{\text{cluster}}
+
\lambda_4 L_{\text{self}} \\
&
+
\lambda_5 L_{\text{CSP}}
+
\lambda_6 L_{\text{DTSE}},
\end{aligned}
\end{equation}

where $L_{\text{rec}}$ denotes the reconstruction loss, $L_{\text{adv}}$ represents the adversarial loss, $L_{\text{cluster}}$ is the clustering objective based on the Student's t-distribution, $L_{\text{CSP}}$ enforces causal consistency, and $L_{\text{DTSE}}$ captures temporal subspace evolution. The hyperparameters $\lambda_1,\ldots,\lambda_6$ control the contribution of each objective during optimization.



\subsection{Optimization and Training Schedule}
\begin{enumerate}
  \item {Generator step.}
  Forward once to obtain $\hat{\mathbf{X}},\, \{\mathbf{z}_t\}_{t=1}^{T},\, \{\mathbf{q}_t\}_{t=1}^{T}$.
  Compute reconstruction and clustering objectives
  $\mathcal{L}_{\mathrm{rec}}$, $\mathrm{KL}(\mathbf{p}\Vert\mathbf{q})$,
  balancing loss, and self-expression loss. Synthesize fake latents and compute
  the adversarial term $\mathcal{L}_{\mathrm{adv}}$. Update encoder/decoder,
  Bi-TGAT, clustering centers, and SE parameters.

  \item {Discriminator step.}
  Sample real latents per cluster (top-$m$ by responsibility) and synthesize fakes.
  Update $\{\mathbf{U}_k\}_{k=1}^{K}$ by minimizing $\mathcal{L}_{\mathrm{D}}$.

  \item {Schedules.}
  Initialize with a larger temperature $\tau$ (softer assignments) and anneal
  $\tau \downarrow$ over training; ramp balancing coefficients; optionally enable
  SE after a warm-up phase to avoid early sparse overfitting.
  \item {Inference and final clustering.}
At test time, we compute $\{\mathbf{z}_t\}$ and $\{\mathbf{q}_t\}$. Final hard labels are $\hat{y}_t = \arg\max_k q_{t,k}$. Optionally, we build an affinity 
\(
\mathbf{A} = |\mathbf{C}| + |\mathbf{C}^\top|
\)
from SE coefficients and apply spectral clustering to refine temporal segments, leveraging the induced block-diagonal structure.
  
\end{enumerate}

\section{Experiment}\label{sec:exp}
All models are executed on AWS cloud environment using 20GB of S3 storage with 30 GB of ml.g4dn.xlarge GPU. The hardware used is a macOS Sonoma version 14.4.1, 16 GB, M1 pro chip. We applied the same python library across all models for homogeneity. We aim to implement our proposed model using python's machine and deep learnings libraries including Keras 2.11 and TensorFlow 2. All the baseline models and proposed models would be tested on Google Colab notebook with 12 GB GPU - A100, High RAM memory support. The hardware we would be using is a macOS ventura version 13.3, 16 GB, M1 pro chip.

\subsection{Dataset and Data Preprocessing} \label{preprocesss}
\begin{figure}[H]
    \centering
    \caption{3D Spatiotemporal View of Surface Pressure (SP)}
    \includegraphics[width=0.45\textwidth]{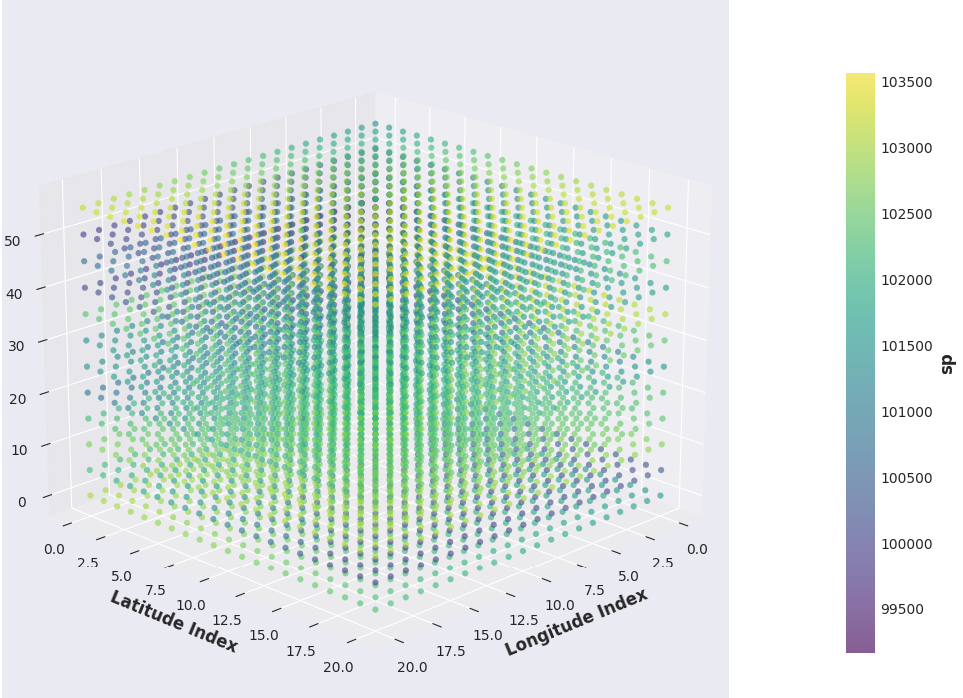}
    \label{fig:kl}
\end{figure}

To ensure generalizability, we experimented with three multivariate spatiotemporal datasets: C3S Arctic Regional Reanalysis (CARRA) dataset~\cite{CARRA}, European Centre for Medium-Range Weather Forecasts (ECMRWF) ERA-5 global reanalysis product~\cite{ERA5}, and daily atmospheric observations~\cite{NCAR}. These datasets are provided alongside our implementation code and publicly available. Figure \ref{fig:kl} illustrates the surface pressure (SP) variable of the ERA5 multivariate spatiotemporal dataset used in the CASC. All datasets follow the same preprocessing steps.
All three data sets consists of daily observations over the course of one year and presented in four dimensions: longitude, latitude, time, and variables. Our proposed model accepts 4D data but to obtain a dimension suitable for our benchmark models, we transform the data from 4D to 2D tabular data \([time, (lon, lat, var)]\)
Existing null values are replaced by the overall mean of the dataset. We apply standard Min-Max Normalization which rescales all features to fall within the range of \([0,1].\)
\vspace{-0.9em}
\vspace{-0.5em}
\subsection{Baseline Methods} We compare our proposed model against state-of-the-art deep clustering models. These include (DEC) \cite{xie2016unsupervised}, (DSC) \cite{faruque2023deep}, ClusterGAN \cite{mukherjee2019clustergan}, , (DTC) \cite{sai2018deep} and DASC \cite{zhou2018deep}, respectively. 
Based on the elbow method, we used \(k = 7\) for ERA5 and NCAR datasets and \(k = 5\) for CARRA datasets in our experiments.

\vspace{-0.9em}
\subsection{Evaluation Metrics}\label{sec:Evaluation}

In the absence of ground truth, we evaluate the performance of our proposed model on six internal cluster validation measures: Silhouette Score \cite{shahapure2020cluster}, Davies-Bouldin score (DB) \cite{ros2023pdbi}, Calinski-Harabas score (CH) \cite{wang2019improved}, Average inter-cluster distance (I-CD) \cite{everitt2011cluster}, Average Variance (Variance) \cite{montgomery2010applied} and Average root mean squared error (RMSE) \cite{willmott2005advantages}. These measures seek to balance the \textit{compactness} and the \textit{separation} of formed clusters through minimizing intra-cluster distance and maximizing the inter-cluster distance respectively.

\begin{table*}[h]
\begin{center}
\caption{Performance evaluation of our proposed model: Each selected model is evaluated on all six metrics. Best results are bolded}
\label{tab:my-finalP}
\resizebox{0.9\textwidth}{!}{%
\begin{tabular}{l|cccccc|c|}
\cline{2-8}
 &
  \multicolumn{6}{c|}{\textbf{Baseline Models}} &
  \textbf{Proposed} \\ \hline
\multicolumn{1}{|c|}{\multirow{7}{*}{\begin{tabular}[c]{@{}c@{}}ERA5\\ \\ \\ \\ 7 optimal\\ clusters\end{tabular}}} &
  \multicolumn{1}{c|}{\textbf{Performance}} &
  \multicolumn{1}{c|}{\textbf{ClusterGAN}} &
  \multicolumn{1}{c|}{\textbf{DTC}} &
  \multicolumn{1}{c|}{\textbf{DSC}} &
  \multicolumn{1}{c|}{\textbf{DEC}} &
  \textbf{DASC} &
  \textbf{CASC} \\ \cline{2-8} 
\multicolumn{1}{|c|}{} &
  \multicolumn{1}{c|}{\textbf{Silhouette \(\uparrow\)}} &
  \multicolumn{1}{c|}{-0.0989} &
  \multicolumn{1}{c|}{0.2284} &
  \multicolumn{1}{c|}{0.2903} &
  \multicolumn{1}{c|}{0.2050} &
  0.1355 &
  \textbf{0.3268} \\ \cline{2-8} 
\multicolumn{1}{|c|}{} &
  \multicolumn{1}{c|}{\textbf{DB \(\downarrow\)}} &
  \multicolumn{1}{c|}{17.1624} &
  \multicolumn{1}{c|}{1.8517} &
  \multicolumn{1}{c|}{1.6741} &
  \multicolumn{1}{c|}{1.7515} &
  2.0325 &
  \textbf{1.5009 }\\ \cline{2-8} 
\multicolumn{1}{|c|}{} &
  \multicolumn{1}{c|}{\textbf{CH \(\uparrow\)}} &
  \multicolumn{1}{c|}{1.2348} &
  \multicolumn{1}{c|}{72.3222} &
  \multicolumn{1}{c|}{\textbf{102.2887}} &
  \multicolumn{1}{c|}{99.3082} &
  72.9257 &
  98.8211 \\ \cline{2-8} 
\multicolumn{1}{|c|}{} &
  \multicolumn{1}{c|}{\textbf{RMSE \(\downarrow\)}} &
  \multicolumn{1}{c|}{22.2032} &
  \multicolumn{1}{c|}{15.0820} &
  \multicolumn{1}{c|}{13.6154} &
  \multicolumn{1}{c|}{13.7425} &
  15.0477 &
  \textbf{13.5158 }\\ \cline{2-8} 
\multicolumn{1}{|c|}{} &
  \multicolumn{1}{c|}{\textbf{Variance \(\downarrow\)}} &
  \multicolumn{1}{c|}{0.1064} &
  \multicolumn{1}{c|}{\textit{0.0450}} &
  \multicolumn{1}{c|}{0.1033} &
  \multicolumn{1}{c|}{\textit{0.0450}} &
  0.1039 &
  0.1038 \\ \cline{2-8} 
\multicolumn{1}{|c|}{} &
  \multicolumn{1}{c|}{\textbf{I-CD \(\uparrow\)}} &
  \multicolumn{1}{c|}{4.0315} &
  \multicolumn{1}{c|}{6.4448} &
  \multicolumn{1}{c|}{6.8481} &
  \multicolumn{1}{c|}{6.8093} &
  5.5229 &
  \textbf{7.4839} \\ \hline
\multicolumn{1}{|l|}{} &
  \multicolumn{1}{l|}{} &
  \multicolumn{1}{c|}{} &
  \multicolumn{1}{c|}{} &
  \multicolumn{1}{c|}{} &
  \multicolumn{1}{c|}{} &
   &
   \\ \hline
\multicolumn{1}{|l|}{\multirow{6}{*}{\begin{tabular}[c]{@{}l@{}}CARRA\\ \\ 5 optimal\\ clusters\end{tabular}}} &
  \multicolumn{1}{c|}{\textbf{Silhouette \(\uparrow\)}} &
  \multicolumn{1}{c|}{-0.0753} &
  \multicolumn{1}{c|}{0.0220} &
  \multicolumn{1}{c|}{0.2437} &
  \multicolumn{1}{c|}{0.2027} &
  -0.1059 &
  \textbf{0.2767} \\ \cline{2-8} 
\multicolumn{1}{|l|}{} &
  \multicolumn{1}{c|}{\textbf{DB \(\downarrow\)}} &
  \multicolumn{1}{c|}{7.7668} &
  \multicolumn{1}{c|}{2.2332} &
  \multicolumn{1}{c|}{1.6844} &
  \multicolumn{1}{c|}{1.6781} &
  11.7039 &
  \textbf{1.5089} \\ \cline{2-8} 
\multicolumn{1}{|l|}{} &
  \multicolumn{1}{c|}{\textbf{CH \(\uparrow\)}} &
  \multicolumn{1}{c|}{4.9468} &
  \multicolumn{1}{c|}{55.5673} &
  \multicolumn{1}{c|}{\textbf{78.7826}} &
  \multicolumn{1}{c|}{68.0469} &
  17.4001 &
  69.7729 \\ \cline{2-8} 
\multicolumn{1}{|l|}{} &
  \multicolumn{1}{c|}{\textbf{RMSE \(\downarrow\)}} &
  \multicolumn{1}{c|}{7.9781} &
  \multicolumn{1}{c|}{7.0421} &
  \multicolumn{1}{c|}{5.8789} &
  \multicolumn{1}{c|}{\textbf{5.5029}} &
  11.3033 &
  5.5424 \\ \cline{2-8} 
\multicolumn{1}{|l|}{} &
  \multicolumn{1}{c|}{\textbf{Variance \(\downarrow\)}} &
  \multicolumn{1}{c|}{0.0021} &
  \multicolumn{1}{c|}{0.0016} &
  \multicolumn{1}{c|}{\textbf{0.0011}} &
  \multicolumn{1}{c|}{0.0160} &
  0.0011 &
  0.0105 \\ \cline{2-8} 
\multicolumn{1}{|l|}{} &
  \multicolumn{1}{c|}{\textbf{I-CD \(\uparrow\)}} &
  \multicolumn{1}{c|}{2.1073} &
  \multicolumn{1}{c|}{2.3191} &
  \multicolumn{1}{c|}{2.5712} &
  \multicolumn{1}{c|}{2.8264} &
  \textbf{3.1404} &
  3.0912 \\ \hline
\multicolumn{1}{|l|}{} &
  \multicolumn{1}{l|}{} &
  \multicolumn{1}{l|}{} &
  \multicolumn{1}{l|}{} &
  \multicolumn{1}{l|}{} &
  \multicolumn{1}{l|}{} &
  \multicolumn{1}{l|}{} &
  \multicolumn{1}{l|}{} \\ \hline
\multicolumn{1}{|l|}{\multirow{6}{*}{\begin{tabular}[c]{@{}l@{}}NCAR\\ Reanalysis 1\\ \\ 7 optimal\\ clusters\end{tabular}}} &
  \multicolumn{1}{c|}{\textbf{Silhouette \(\uparrow\)}} &
  \multicolumn{1}{c|}{-0.2659} &
  \multicolumn{1}{c|}{0.6230} &
  \multicolumn{1}{c|}{0.61563} &
  \multicolumn{1}{c|}{0.6454} &
  0.1132 &
  \textbf{0.6541 }\\ \cline{2-8} 
\multicolumn{1}{|l|}{} &
  \multicolumn{1}{c|}{\textbf{DB \(\downarrow\)}} &
  \multicolumn{1}{c|}{9.3799} &
  \multicolumn{1}{c|}{\textbf{0.7570}} &
  \multicolumn{1}{c|}{0.7804} &
  \multicolumn{1}{c|}{0.7639} &
  2.0879 &
  0.7612 \\ \cline{2-8} 
\multicolumn{1}{|l|}{} &
  \multicolumn{1}{c|}{\textbf{CH \(\uparrow\)}} &
  \multicolumn{1}{c|}{3.3104} &
  \multicolumn{1}{c|}{864.1750} &
  \multicolumn{1}{c|}{862.3665} &
  \multicolumn{1}{c|}{839.4187} &
  192.6249 &
  \textbf{868.7555} \\ \cline{2-8} 
\multicolumn{1}{|l|}{} &
  \multicolumn{1}{c|}{\textbf{RMSE \(\downarrow\)}} &
  \multicolumn{1}{c|}{12.1719} &
  \multicolumn{1}{c|}{3.1380} &
  \multicolumn{1}{c|}{3.1410} &
  \multicolumn{1}{c|}{3.1809} &
  6.0048 &
  3.1180 \\ \cline{2-8} 
\multicolumn{1}{|l|}{} &
  \multicolumn{1}{c|}{\textbf{Variance \(\downarrow\)}} &
  \multicolumn{1}{c|}{0.1770} &
  \multicolumn{1}{c|}{0.1770} &
  \multicolumn{1}{c|}{0.1770} &
  \multicolumn{1}{c|}{0.1770} &
  0.1770 &
  0.1770 \\ \cline{2-8} 
\multicolumn{1}{|l|}{} &
  \multicolumn{1}{c|}{\textbf{I-CD \(\uparrow\)}} &
  \multicolumn{1}{c|}{0.7357} &
  \multicolumn{1}{c|}{0.8603} &
  \multicolumn{1}{c|}{0.8745} &
  \multicolumn{1}{c|}{\textbf{0.9465}} &
  4.0097 &
  0.9098 \\ \hline
\end{tabular}%
}
\end{center}
\end{table*}

\subsection{Experimental Results and Discussion} \label{tab:Results} Table~\ref{tab:my-finalP} summarizes the clustering performance of all competing approaches across the ERA5, CARRA, and NCAR Reanalysis 1 datasets using six internal validation metrics. Overall, the proposed CASC framework consistently demonstrates superior clustering quality and robustness across the three datasets, achieving either the best or highly competitive performance on most evaluation measures. These results indicate that incorporating attention-guided adversarial learning within a temporal subspace clustering framework enables the model to learn more discriminative latent representations and uncover meaningful temporal structures from complex multivariate climate data. For the ERA5 dataset, CASC achieved the highest Silhouette score (0.3268), the lowest Davies--Bouldin (DB) index (1.5009), the lowest RMSE (13.5158), and the largest Inter-Cluster Distance (I-CD) value (7.4839). These results suggest that the clusters produced by CASC are simultaneously more compact and better separated than those generated by the competing methods. Although DSC obtained the highest Calinski--Harabasz (CH) score (102.2887), the difference relative to CASC (98.8211) is marginal. The superior performance of CASC can likely be attributed to the attention mechanism, which selectively emphasizes informative temporal patterns while suppressing noisy or redundant climate signals. Furthermore, the adversarial learning component encourages the latent representation to capture the underlying data distribution more effectively, resulting in improved cluster separability. In contrast, ClusterGAN exhibits poor performance across all metrics, including a negative Silhouette score and an extremely high DB index. This behavior suggests that adversarial generation alone is insufficient for capturing the intrinsic temporal subspace structure present in climate reanalysis data. Similarly, DASC underperforms relative to CASC, indicating that adversarial learning benefits substantially from the addition of attention-guided feature selection. The CARRA dataset presents a more challenging clustering scenario due to its higher variability and localized regional characteristics. Nevertheless, CASC again achieved the best Silhouette score (0.2767) and the lowest DB index (1.5089), indicating improved cluster cohesion and separation. Although DEC produced the lowest RMSE (5.5029) and DSC achieved the highest CH score (78.7826), the differences relative to CASC remain small. Interestingly, DASC obtained the highest I-CD value (3.1404), marginally outperforming CASC (3.0912). This observation suggests that while DASC generates clusters with slightly larger average inter-cluster distances, CASC achieves a better balance between intra-cluster compactness and inter-cluster separation, which is reflected in its superior Silhouette and DB scores. The comparatively weak performance of ClusterGAN and DTC may be due to their inability to model nonlinear temporal dependencies adequately, whereas CASC benefits from attention-driven representation learning that can adaptively focus on dominant climatic patterns. The NCAR Reanalysis 1 dataset exhibits the clearest cluster structure among the three datasets, as evidenced by the substantially higher Silhouette and CH scores achieved by most methods. On this dataset, CASC attained the highest Silhouette score (0.6541), the highest CH score (868.7555), and the lowest RMSE (3.1180).
\vspace{-0.9em}
\begin{figure}[H]
    \centering
    \caption{CASC Clustering Visualization}
    \includegraphics[width=0.5\textwidth]{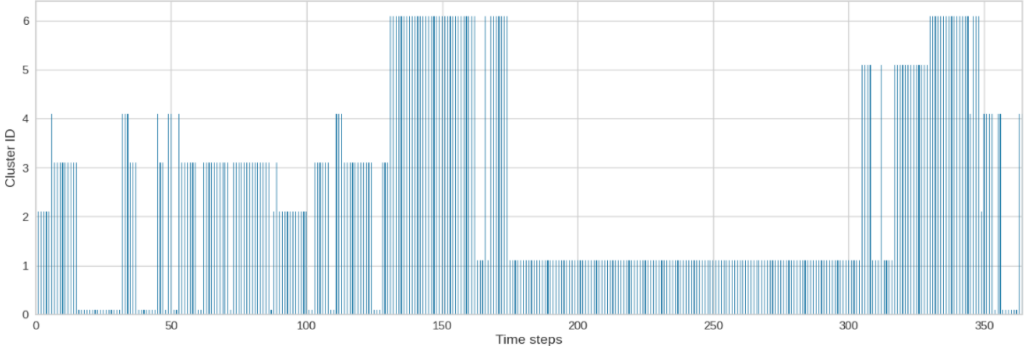}
    \label{fig:j}
\end{figure}
\vspace{-0.9em}
Figure \ref{fig:j} illustrates CASC results from clustering along temporal dimension spanning 365  daily observations, where each bar represents the cluster assignment of a single day. The model identifies seven distinct temporal regimes (Cluster IDs 0–6), revealing clear transitions in the underlying spatiotemporal dynamics throughout the year. During the first half of the year, the cluster assignments fluctuate frequently among Clusters 2–6, indicating periods of rapidly changing environmental conditions and diverse temporal patterns. A notable and prolonged dominance of Cluster 1 is observed between approximately days 175 and 305, suggesting the presence of a highly stable and persistent spatiotemporal regime during this period. Following this stable phase, the clustering structure becomes more heterogeneous, with frequent transitions among Clusters 4, 5, and 6 near the end of the year, indicating increased variability and regime shifts. The uneven distribution of cluster memberships demonstrates that some temporal states occur more frequently than others, reflecting the ability of CASC to capture both persistent and transient patterns in multivariate spatiotemporal data.
Although DTC achieved a marginally lower DB index (0.7570 versus 0.7612), the difference is negligible and does not alter the overall ranking. These findings indicate that CASC successfully identifies highly coherent temporal regimes while maintaining strong separation between clusters. Interestingly, DASC achieved a substantially larger I-CD value (4.0097) than all other methods.
While a larger inter-cluster distance generally suggests improved separation, it may also indicate the formation of overly dispersed clusters with reduced internal consistency. This interpretation is supported by DASC's lower Silhouette score and higher RMSE compared with CASC. Consequently, the results suggest that maximizing inter-cluster distance alone does not necessarily yield optimal clustering quality. Instead, effective clustering requires balancing both compactness and separation, which CASC appears to achieve more consistently. Across all datasets, several important trends emerge. First, deep clustering approaches such as DSC, DEC, and CASC consistently outperform traditional adversarial clustering methods, highlighting the importance of learning structured latent representations for climate data. Second, attention-guided mechanisms provide measurable benefits by enabling the model to focus on the most informative temporal features while reducing the influence of noise and irrelevant variability. Third, adversarial learning appears most effective when combined with attention-enhanced representation learning, as demonstrated by the substantial performance gap between DASC and CASC. Finally, the results reveal that no single metric fully captures clustering quality. For example, DSC occasionally achieves the highest CH score, while DASC obtains the largest I-CD value on certain datasets. However, CASC consistently achieves strong performance across multiple complementary metrics, indicating a more balanced and reliable clustering solution. Overall, the experimental results demonstrate that CASC effectively captures the complex nonlinear temporal relationships inherent in multivariate climate reanalysis datasets. Its ability to jointly optimize representation learning, temporal attention, and adversarial subspace discovery leads to more compact, well-separated, and interpretable clusters compared with existing state-of-the-art baseline methods.
\vspace{-0.5em}
\subsection{Ablation Study}
\begin{table}[ht]
\caption{Ablation Study}
\label{tab:my-table8}
\resizebox{\columnwidth}{!}{%
\begin{tabular}{|ccccccc|}
\hline
\multicolumn{7}{|c|}{Performance - based} \\ \hline
\multicolumn{1}{|c|}{\textbf{}} &
  \multicolumn{1}{c|}{Silhouette \(\uparrow\) } &
  \multicolumn{1}{c|}{DB \(\downarrow\)} &
  \multicolumn{1}{c|}{CH \(\uparrow\) } &
  \multicolumn{1}{c|}{RMSE \(\downarrow\)} &
  \multicolumn{1}{c|}{Variance \(\downarrow\)} &
  ICD \(\uparrow\)  \\ \hline
\multicolumn{1}{|c|}{CASC$_{CSP}$} &
  \multicolumn{1}{c|}{0.2124} &
  \multicolumn{1}{c|}{2.1486} &
  \multicolumn{1}{c|}{84.6793} &
  \multicolumn{1}{c|}{{14.4220}} &
  \multicolumn{1}{c|}{0.1038} &
   6.1603\\ \hline
\multicolumn{1}{|c|}{CASC$_{DTSE}$} &
  \multicolumn{1}{c|}{0.1965} &
  \multicolumn{1}{c|}{1.9576} &
  \multicolumn{1}{c|}{{91.9739}} &
  \multicolumn{1}{c|}{14.0709} &
  \multicolumn{1}{c|}{\textbf{0.1032}} &
  {5.6349} \\ \hline
\multicolumn{1}{|c|}{CASC$_{SETD}$} &
  \multicolumn{1}{c|}{{0.2827}} &
  \multicolumn{1}{c|}{{1.6941}} &
  \multicolumn{1}{c|}{\textbf{99.9202}} &
  \multicolumn{1}{c|}{13.716} &
  \multicolumn{1}{c|}{{0.1035}} &
  {6.2759} \\ \hline
\multicolumn{1}{|c|}{CASC} &
  \multicolumn{1}{c|}{\textbf{0.3268}} &
  \multicolumn{1}{c|}{\textbf{1.5009}} &
  \multicolumn{1}{c|}{98.8211} &
  \multicolumn{1}{c|}{\textbf{13.5158}} &
  \multicolumn{1}{c|}{{0.1038}} &
  \textbf{7.4839} \\ \hline
\end{tabular}%
}
\end{table}
To evaluate the contribution of each proposed component, we performed an ablation study on the ERA5 dataset by independently analyzing the effects of the Causal Subspace Preservation (CSP) loss, Dynamic Temporal Subspace Evolution (DTSE) loss, and the Subspace-Aware Energy-based Temporal Discriminator (SETD). As shown in Table~\ref{tab:my-table8}, each module contributes complementary benefits to the clustering framework. The CSP-only variant improves cluster compactness and separation by preserving causally consistent latent subspaces, achieving a silhouette score of \textit{0.2124} and an ICD of \textit{6.1603}. However, its relatively high DB index and RMSE indicate limited capability in modeling complex temporal dynamics. The DTSE-only variant produces the lowest variance \textit{0.1032}, demonstrating its effectiveness in enforcing temporal smoothness and cluster stability, although its cluster discrimination remains weaker than the other variants. Among the individual components, CASC$_{SETD}$ achieves the strongest performance, obtaining a silhouette score of \textit{0.2827}, a DB index of \textit{1.6941}, and the highest CH score of \textit{99.9202}. These results highlight the effectiveness of SETD in learning highly discriminative latent representations through energy-based subspace modeling, which improves cluster separability and reduces overlap between latent subspaces. The complete CASC framework achieves the best overall performance, with the highest silhouette score \textit{0.3268}, lowest DB index \textit{1.5009}, lowest RMSE \textit{13.5158}, and largest ICD \textit{7.4839}. These results demonstrate that the joint integration of CSP, DTSE, and SETD effectively combines causal consistency, temporal coherence, and subspace discrimination. 



\section{Conclusion} \label{concl}

In this paper, we introduced \textit{Causal Adversarial Subspace Clustering (CASC)}, a novel end-to-end deep clustering framework designed for high-dimensional multivariate spatiotemporal data. CASC integrates attention-guided spatiotemporal representation learning, graph-based self-expressive subspace clustering, causal subspace preservation, dynamic temporal subspace evolution, and a Subspace-Aware Energy-based Temporal Discriminator (SETD) within a unified adversarial learning framework. By jointly modeling spatial dependencies, temporal dynamics, causal relationships, and latent subspace structures, the proposed model learns discriminative and temporally coherent representations that effectively capture the complex nonlinear characteristics of real-world spatiotemporal phenomena. Experimental results on multiple climate and environmental datasets demonstrate that CASC consistently produces compact, well-separated, and stable clusters, outperforming several state-of-the-art deep clustering baselines across a wide range of internal clustering evaluation metrics.

The proposed framework highlights the benefits of combining causal representation learning with adversarial subspace clustering to improve both clustering quality and latent-space interpretability. The ablation analysis further reveals that the Causal Subspace Preservation (CSP) loss, Dynamic Temporal Subspace Evolution (DTSE) loss, and SETD module contribute complementary strengths, with their joint optimization yielding the strongest overall performance. Beyond climate analytics, CASC provides a general framework for unsupervised pattern discovery in complex spatiotemporal domains such as remote sensing, environmental monitoring, transportation systems, healthcare analytics, and smart-city applications. Future research will investigate adaptive subspace discovery, dynamic causal graph learning, contrastive representation learning, uncertainty-aware clustering, and physics-informed constraints to further improve scalability, robustness, and interpretability in large-scale spatiotemporal data mining applications.

\section*{Acknowledgment}

The authors acknowledge the minimal use of OpenAI ChatGPT-4o to refine the components within Figure 1 for the sole purpose to improve clarity and presentation.

\bibliography{biblo}
\bibliographystyle{ieeetr}


\end{document}